%% file: main.tex
\documentclass[10pt, twocolumn, letterpaper, dvipsnames, table]{article}

\usepackage[accsupp]{axessibility}
\usepackage[pagenumbers]{iccv} 
\usepackage{bm}
\usepackage{tabularx}
\usepackage{array}
\usepackage{multirow}
\usepackage{booktabs}
\usepackage{natbib}
\input{macro.tex}

\input{define.tex}
\input{preamble}

\definecolor{iccvblue}{rgb}{0.21,0.49,0.74}
\usepackage[pagebackref, breaklinks, colorlinks, allcolors=iccvblue]{hyperref}

\def\paperID{5325} 
\def\confName{ICCV}
\def\confYear{2025}

\title{LONG3R: Long Sequence Streaming 3D Reconstruction}

\author{
Zhuoguang Chen$^{1,2*}$ \quad
Minghui Qin$^{2*}$ \quad
Tianyuan Yuan$^{2,3*}$ \quad
Zhe Liu$^2$ \quad
Hang Zhao$^{2,1,3\dagger}$
\\[0.5em]
$^1$Shanghai Artificial Intelligence Laboratory \ \ \ 
$^2$IIIS, Tsinghua University \ \ \
$^3$Shanghai Qi Zhi Institute 
\vspace{-1em}
}

\begin{document}
    \maketitle
    {\let\thefootnote\relax\footnotetext{{$^*$ Equal contribution.} $^\dagger$Corresponding author.}}
    \input{sec/0_abstract}
    \input{sec/1_intro}
    \input{sec/2_relativework}
    \input{sec/3_method}
    \input{sec/4_experiment}

    \input{sec/5_conclusion}
    \input{sec/6_acknowledgment}
    { \small \bibliographystyle{ieeenat_fullname} \bibliography{main} }

\end{document}

%% file: macro.tex
\newcommand{\It}{\bm{I}_t}

\newcommand{\ft}{\bm{F}^I_t}

%% file: define.tex
\makeatletter
\renewcommand{\paragraph}{\@startsection{paragraph}{4}{\z@}{1.2ex plus 0.5ex minus .2ex}{-1em}{\normalsize\bf}}

\makeatother

%% file: preamble.tex
\definecolor{color1}{HTML}{00FF00}
\definecolor{color2}{HTML}{33FF00}
\definecolor{color3}{HTML}{66FF00}
\definecolor{color4}{HTML}{99FF00}
\definecolor{color5}{HTML}{CCFF00}
\definecolor{color6}{HTML}{FFCC00}
\definecolor{color7}{HTML}{FF9900}
\definecolor{color8}{HTML}{FF0000}

\newcommand{\ourwork}{LONG3R\xspace}

%% file: sec/0_abstract.tex
\begin{abstract}
Recent advancements in multi-view scene reconstruction have been significant, yet existing methods face limitations when processing streams of input images. These methods either rely on time-consuming offline optimization or are restricted to shorter sequences, hindering their applicability in real-time scenarios. In this work, we propose LONG3R (\textbf{LO}ng sequence streami\textbf{NG} \textbf{3}D \textbf{R}econstruction), a novel model designed for streaming multi-view 3D scene reconstruction over longer sequences. Our model achieves real-time processing by operating recurrently, maintaining and updating memory with each new observation. 
We first employ a memory gating mechanism to filter relevant memory, which, together with a new observation, is fed into a dual-source refined decoder for coarse-to-fine interaction.
To effectively capture long-sequence memory, we propose a 3D spatio-temporal memory that dynamically prunes redundant spatial information while adaptively adjusting resolution along the scene. To enhance our model’s performance on long sequences while maintaining training efficiency, we employ a two-stage curriculum training strategy, each stage targeting specific capabilities. Experiments demonstrate that LONG3R outperforms state-of-the-art streaming methods, particularly for longer sequences, while maintaining real-time inference speed. 
Project page: https://zgchen33.github.io/LONG3R/.

\end{abstract}

%% file: sec/1_intro.tex
\section{Introduction}
\label{sec:intro}

\input{fig/teaser.tex}

Recovering dense geometry from a sequence of images is a fundamental task in 3D computer vision. It has widespread applications in robotics, autonomous driving, and indoor and outdoor scene reconstruction. 
Traditional approaches typically address this task through various methods, including Structure from Motion (SfM)~\cite{crandall2011discrete,aanaes2016large,schonberger2016structure,snavely2006photo,sweeney2015optimizing,wilson2014robust,wu2013towards}, keypoint detection~\cite{rublee2011orb,lowe2004distinctive,lowe1999object}, bundle adjustment~\cite{wu2011multicore,triggs2000bundle,agarwal2010bundle}, Simultaneous Localization and Mapping (SLAM)~\cite{newcombe2011dtam,klein2007parallel,davison2007monoslam}, and Multi-View Stereo~\cite{newcombe2011dtam,schonberger2016pixelwise,galliani2015massively,furukawa2009accurate}. While these methods have achieved notable success, they rely on hand-crafted heuristics, requiring significant engineering effort when assembled into a pipeline~\cite{schoenberger2016sfm, schonberger2016pixelwise}.

Recently, a new class of methods, beginning with DUSt3R~\cite{wang2024dust3r} and MASt3R~\cite{mast3r_leroy2024grounding}, has tackled this problem using end-to-end neural networks. These models directly regress 3D representations (\ie, pointmaps) from image pairs, offering a simple yet highly generalizable approach that has quickly gained traction in the community. A particularly promising direction extends this paradigm to online processing of streaming input images. For instance, Spann3R~\cite{wang2024spann3r} introduced a recurrent model with memory to process streaming input images in real time, enabling various practical applications. However, despite its efficiency, Spann3R struggles with long input sequences due to three key issues: (1) its memory is only attended once per iteration, preventing effective reuse, (2) its memory becomes spatially redundant as images accumulate, and (3) its training strategy does not support adaptation to long sequences.

To address the aforementioned challenges, in this paper, we propose \ourwork (\textbf{LO}ng sequence streami\textbf{NG} \textbf{3}D \textbf{R}econstruction), a novel model designed for streaming multi-view 3D scene reconstruction over longer sequences, as illustrated in \cref{fig:teaser}. 
We define long-sequence reconstruction as real-time processing of tens to hundreds of frames with near-constant memory requirements.
Like Spann3R~\cite{wang2024spann3r}, our approach employs a recurrent network with a spatio-temporal 3D memory bank to process streaming image sequences. Given a new observation, our model retrieves relevant memories, interacts with the current view to predict its pointmap, and updates the memory accordingly. To enhance long-sequence processing, we introduce three key innovations:
\begin{enumerate}
    \item Memory Gating \& Dual-Source Decoder: We introduce a memory gating mechanism that selectively retains memories relevant to the current observation, followed by a Dual-Source Refined Decoder that enables coarse-to-fine interaction between observations and memories.
    \item 3D Spatio-temporal Memory: We propose a dynamic 3D memory module that automatically prunes redundant memories and adapts resolution to the scene scale, balancing memory efficiency and reconstruction accuracy.
    \item Two-stage Curriculum Training: We adopt a two-stage curriculum training strategy that progressively increases sequence length, enhancing the model’s ability to handle increasingly complex memory interactions.
\end{enumerate}
We conduct extensive experiments on multiple 3D datasets, comparing \ourwork with various state-of-the-art methods. Results demonstrate that our approach significantly improves long-sequence streaming reconstruction while maintaining real-time processing capability.

%% file: fig/teaser.tex
\begin{figure}[t]
    \centering
    \includegraphics[width=\linewidth]{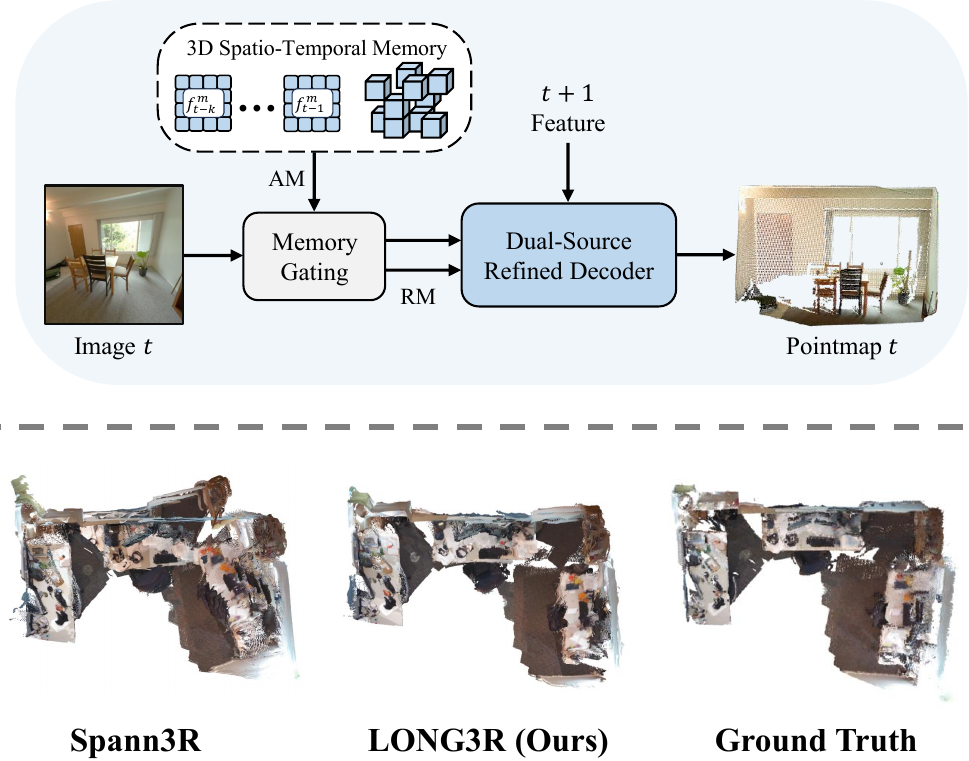}
    \caption{(Top): Overview of our \ourwork framework, which integrates a 3D spatio-temporal memory module and memory gating to refine features through a dual-source refined decoder. Here, \textbf{AM} denotes \textit{All Memory} and \textbf{RM} denotes \textit{Relevant Memory}. (Bottom): Qualitative comparison of 3D reconstructions from Spann3R, our method \ourwork, and the ground truth. \ourwork can achieve more accurate point prediction.}
    \label{fig:teaser}
\end{figure}

%% file: sec/2_relativework.tex
\section{Related Work}
\input{fig/methods_fig}
\paragraph{Traditional 3D Reconstruction.}
3D reconstruction from visual data modalities has been a long-standing research challenge
in computer vision and graphics. This field has evolved through paradigm shifts
from Structure-from-Motion (SfM) frameworks~\cite{cadena2016past,durrant2006simultaneous}
and SLAM systems~\cite{davison2007monoslam,mur2015orb,engel2014lsd,klein2007parallel,newcombe2011dtam}
to advanced scene representation methods like Neural Radiance Fields (NeRF)~\cite{mildenhall2021nerf,muller2022instant,chen2022tensorf,fridovich2022plenoxels,wang2021neus, ye2024blending}
and 3D Gaussian Splatting~\cite{kerbl20233d,huang20242dgs,yu2024mip, zhang2024drone, yang2024spectrallyprunedgaussianfields}. Contemporary
reconstruction frameworks expose fundamental limitations of classical geometric approaches
when confronted with sparse observations, ill-posed problems, or long unconstrained
sequences. Traditional methods based on geometric optimization—such as explicit
feature matching and iterative bundle adjustment—are inherently sensitive to
observation sparsity, computationally redundant, and notably slow due to the
absence of learned priors for regularizing underconstrained scenarios. In contrast,
our streaming 3D reconstruction method, \ourwork, directly regresses 3D
pointmaps from images using learned priors, resulting in significantly faster reconstruction.

\paragraph{Learning-Based 3D Reconstruction.}
Replacing traditional handcrafted components with learning-based approaches, including
learning-based priors~\cite{sarlin2020superglue,detone2018superpoint,sun2021loftr,dusmanu2019d2,teed2021droid,chen2024leap,yu2024mip,yang2020d3vo,zhu2024nicer,tateno2017cnn},
depth estimation~\cite{piccinelli2024unidepth,hu2024metric3d,li2018megadepth,yang2024depth,godard2019digging,bhat2023zoedepth,piccinelli2024unidepth,ranftl2020towards,ranftl2021vision},
and end-to-end system optimization~\cite{wang2024vggsfm,teed2021droid,tang2018ba,yao2018mvsnet},
is becoming a prevailing trend for scalable scene representation. This trend elevates
data-driven methodologies to prominence within 3D reconstruction frameworks,
with pointmap representations~\cite{fei2024driv3r,lu2024lora3d,zhang2025flare,lu2024align3r,Yang_2025_Fast3R,dong2024reloc3r,leroy2024mast3r,wang2024dust3r,wang2024spann3r,wang2025cut3r,liu2024slam3r}
catalyzing the evolution of this field. Drawing inspiration from the CroCo~\cite{croco,croco_v2}
cross-view completion paradigm, DUSt3R~\cite{wang2024dust3r} pioneers a geometry-agnostic
pointmap prediction mechanism without prior calibration. MASt3R~\cite{mast3r_leroy2024grounding}
implements a coarse-to-fine feature matching strategy, which improves the
predictions of the metric-scale point map. Metric scale dense pointmaps predictions~\cite{mast3r_leroy2024grounding} can serve as a front-end to improve
initialization and triangulation processes in SFM/SLAM pipelines~\cite{duisterhof2024mast3r_sfm,murai2024mast3r_slam}.
SLAM3R~\cite{liu2024slam3r} and Reloc3r~\cite{dong2024reloc3r} propose to leverage
a feedforward network for initial mapping and localization estimation, instead
of relying on computationally intensive bundle adjustment-based backends. MV-DUSt3R~\cite{tang2024mvdust3r} incorporates multi-view decoder blocks for cross-view
information exchange and cross-reference view blocks to enhance robustness against
reference view selection. For dynamic scene reconstruction, both MonST3R~\cite{zhang2024monst3r}
and CUT3R~\cite{wang2025cut3r} reconstruct temporally coherent 3D representations
from unconstrained monocular video sequences.

\paragraph{Streaming Reconstruction.}

Current streaming scene reconstruction methods using pairwise image matching struggle with long sequences. Accumulative scale drift and error propagation in pose estimation degrade 3D reconstruction consistency. Traditional monocular SLAM systems~\cite{teed2021droid,zhu2024nicer,engel2014lsd,forster2016svo} mitigate these issues through optimized tracking, optical flow-integrated bundle adjustment, and loop closure detection. Their reliance on predefined intrinsic parameters limits generalizability, and reconstruction quality heavily depends on depth estimation accuracy. Spann3R~\cite{wang2024spann3r} refines point map predictions using a hybrid memory feature bank with attention interaction, enabling feedforward scene reconstruction without iterative optimization. CUT3R~\cite{wang2025cut3r} uses persistent state tokens with transformer-based recurrent
state updates for online reconstruction from streaming sequences, yet
suffers from limited extreme viewpoint extrapolation capabilities due to
deterministic inference and potential drift accumulation in extended sequences
lacking global alignment. Our model utilizes spatio-temporal contextual information
during training and inference phases to reduce cumulative
errors caused by the lack of loop closure detection and post-optimization.

%% file: fig/methods_fig.tex
\begin{figure*}[t]
    \centering
    \begin{minipage}{0.7\textwidth}  
        \centering
        \includegraphics[width=\textwidth]{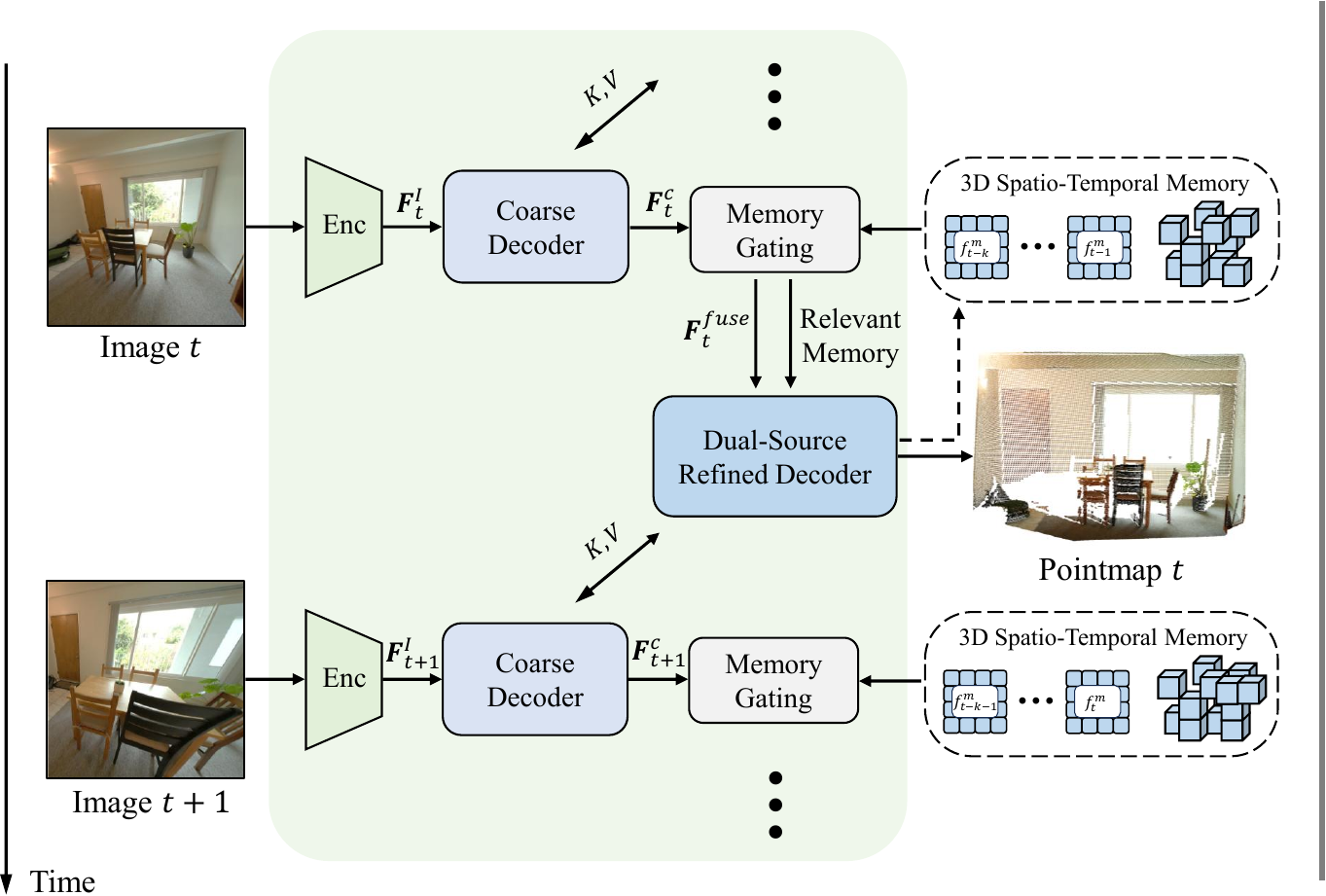}
        \subcaption{Main Architecture.}
        \label{fig:method_a}
    \end{minipage}
    \begin{minipage}{0.29\textwidth}  
        \centering
        \begin{minipage}{\textwidth} 
            \centering
            \includegraphics[width=\textwidth]{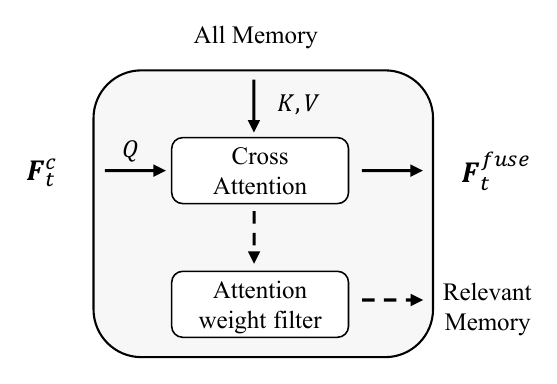}
            \subcaption{Attention-based Memory Gating.}
            \label{fig:method_b}
            \vspace{2em}
        \end{minipage}
        \vfill
        \begin{minipage}{\textwidth} 
            \centering
            \includegraphics[width=\textwidth]{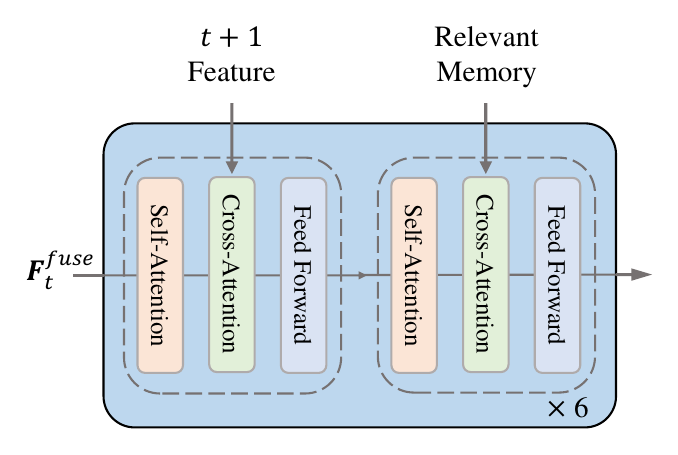}
            \subcaption{Dual-Source Refined Decoder.}
            \label{fig:method_c}
        \end{minipage}
    \end{minipage}
    \caption{\textbf{Method Overview}. (a) 
    Illustrates the overall architecture, where image features $F^I_t$ first interact with $F^I_{t-1}$ in the Coarse Decoder to generate $F^c_t$, after which a memory-gating module filters irrelevant entries from the spatio-temporal memory $F_{\text{mem}}$. The Dual-Source Refined Decoder subsequently interacts with both the filtered memory and features from $t+1$, ultimately generating the pointmap $t$.
    (b) Details the attention-based memory gating module, which selects relevant information from the memory. (c) Illustrates the dual-source refined decoder, which alternately interacts with the next-frame features and relevant memory features through multiple self- and cross-attention layers to optimize memory information utilization and maintain alignment with the subsequent frame.}
    \label{fig:all_method}
\end{figure*}

%% file: sec/3_method.tex
\section{Method}

Our proposed streaming 3D reconstruction method, \ourwork, is depicted in ~\cref{fig:all_method}. It starts by encoding features from consecutive frames, which are then processed by a Coarse Decoder to generate a rough 3D structure. A memory gating mechanism then filters the spatio-temporal memory to retain relevant memory. The relevant memory, along with $t + 1$ context, is used by a Dual-Source Refined Decoder to produce accurate reconstructions.

\subsection{Feature Encoding and Coarse Decoding}
\label{method_preliminary}
Following previous studies~\cite{wang2024spann3r,wang2024dust3r}, the input
image $\It$ is processed by a ViT encoder that partitions it into patches and
linearly projects them into visual feature tokens:
\begin{equation}
    \ft = \mathrm{Encoder}(\It).
\end{equation}
These tokens are then forwarded to a Coarse Decoder, implemented as a generic
transformer composed of $B$ PairwiseBlocks. Each PairwiseBlock, which comprises
self-attention, cross-attention, and an MLP, is used as follows:
\begin{equation}
    \bm{F}^{c}_{t, i}= \text{PairwiseBlock}^{c}_{i}\left(\bm{F}^{c}_{t, i-1}, \bm
    {F}^{r}_{t-1, i-1}\right),
\end{equation}
for $i = 1, 2, \dots, B$, with $\bm{F}^{c}_{t, 0}= \ft$. Here, $\bm{F}^{r}_{t-1,
i-1}$ denotes the refined tokens produced by the corresponding blocks of the 
Refined Decoder for the $(t-1)$-th frame. By efficiently interacting with
temporal features from the previous frame, the Coarse Decoder generates coarse
representations that serve as the basis for subsequent processing.

\subsection{Attention-based Memory Gating}
\label{mem_gating} As depicted in \cref{fig:method_b}, the memory gating
mechanism serves two primary functions: aggregating information from all memory entries
and filtering out irrelevant memory tokens to reduce the computational load of the
subsequent Refined Decoder. For brevity, we denote the final PairwiseBlock output
$\bm{F}^{c}_{t,B}$ by $\bm{F}^{c}_{t}$, which then attends to the memory keys
$\bm{F}^{K}_\text{mem}$ and values $\bm{F}^{V}_\text{mem}$ via cross-attention:

\begin{equation}
    W_{t}= \mathrm{Softmax}\!\left(\frac{\bm{F}^{c}_{t}\left(\bm{F}^{K}_\text{mem}\right)^{\top}}{\sqrt{C}}
    \right), \label{eq:attn_weight}
\end{equation}

\begin{equation}
    \bm{F}_{t}^\text{fuse}= W_{t}\bm{F}^{V}_\text{mem},
\end{equation}

\noindent
where $W_{t}\in \mathbb{R}^{P\times S}$ denotes the attention weights for each
token in the current query relative to all memory keys, with $P$ as the number of
tokens in the current frame and $S$ as the number of memory tokens.

To filter out memory irrelevant to the current observation, we employ the attention weights
$W_{t}\in \mathbb{R}^{P \times S}$ and an attention threshold $\tau$ ($5 \times 10
^{-4}$). Specifically, for each memory index $s \in \{1, \ldots, S\}$, if there exists
at least one token $p \in \{1, \ldots, P\}$ such that $W_{t}(p, s) > \tau$, the
$s$-th memory feature is retained; otherwise, it is discarded. Thus, the
relevant memory $\bm{F}_\text{r\_mem}$ is given by:
\begin{equation}
    \delta(s) =
    \begin{cases}
        1, & \text{if }\max_{p}W_{t}(p, s) > \tau, \\
        0, & \text{otherwise.}
    \end{cases}
\end{equation}
\begin{equation}
    \bm{F}_\text{r\_mem}= \{\, \bm{F}_\text{mem}(s) \mid \delta(s) = 1 \,\}.
\end{equation}
Here, $\delta(s)$ is an indicator function. Notably, $\bm{F}_\text{mem}$ comprises
two components, $\bm{F}^{K}_\text{mem}$ and $\bm{F}^{V}_\text{mem}$, both of which undergo
the same above filtering process. This mechanism ensures that only memory elements
with sufficient relevance, as determined by the attention weights, contribute to
the following refining process.

\subsection{Dual-Source Refined Decoder}
\label{refined_decoder} To maximize the utilization of memory information and maintain
alignment with the subsequent frame, we propose the Dual-Source Refined Decoder,
as illustrated in ~\cref{fig:method_c}. Unlike the Coarse Decoder, which
consists solely of generic PairwiseBlocks, the Dual-Source Refined Decoder alternates
between two types of blocks: a PairwiseBlock and a MemoryBlock. This design allows
the current-frame tokens to fully exploit and aggregate both spatio-temporal memory
tokens and features from the next frame.

Let $\bm{F}^{r}_{t, i}$ denote the refined feature representation at block $i$.
The overall operation can be formulated as:

\begin{equation}
    \bm{F}^{r}_{t,i}=
    \begin{cases}
        \mathrm{PairwiseBlock}(\bm{F}^{r}_{t,i-1}, \bm{F}^{c}_{t+1,i-1}), & i \text{ odd},  \\
        \mathrm{MemoryBlock}(\bm{F}^{r}_{t,i-1}, \bm{F}_\text{r\_mem}),      & i \text{ even},
    \end{cases}
\end{equation}
for $i = 1, 2, \dots, B$, with $\bm{F}^{r}_{t, 0}=\bm{F}_{t}^\text{fuse}$. The
PairwiseBlock (applied for odd $i$) facilitates feature interactions between the
refined current-frame features and the coarse tokens from the next frame, while the
MemoryBlock (applied for even $i$) integrates these refined features with the relevant
memory tokens $\bm{F}_\text{r\_mem}$, thereby enhancing long-range spatio-temporal
dependencies. This alternating structure enables the decoder to construct a robust,
context-aware feature representation by leveraging both immediate and historical
information.

Following the decoder, an explicit 3D reconstruction prediction is generated from
its outputs using a DPT head.

\subsection{3D Spatio-Temporal Memory}
\label{3dmem}

Our memory mechanism handles long sequences by concurrently
maintaining short-term temporal memory and long-term 3D spatial memory. 
The memory consists of historical tokens generated by the Dual-Source Refined Decoder.
With a fixed storage capacity, our long-term 3D spatial memory maintains an overall spatial
representation while avoiding redundant tokens. This memory design efficiently
captures essential spatio-temporal features without overwhelming memory resources. 

The short-term temporal memory stores historical tokens from the time window $[t
-K, t-1]$, where $K$ denotes the window length. It stores key features $f^{K}\in
\mathbb{R}^{(K \cdot P) \times C}$ and value features $f^{V}\in \mathbb{R}^{(K \cdot
P) \times C}$, ensuring the effective utilization of time-dependent information.

The long-term 3D spatial memory mitigates GPU memory constraints and improves
inference speed by managing tokens within $[1, t-K]$ while limiting their
number. Inspired by the occupancy mechanism, we employ voxels as
storage units, with each voxel retaining a single token. This sparsification
balances the number of memory units with the scene’s spatial size. However,
since different scenes require varying voxel sizes and the model’s optimization
is metric-invariant, predefined voxel sizes are unsuitable. To address this, we introduce
an adaptive voxel size strategy.

\input{fig/3dmem_prune}

\input{table/7scenes_table}

\input{fig/7scenes_vis}
\paragraph{Adaptive Voxel Size.}
Since the memory stores patch-based tokens, we first compute a unique 3D position
$P$ for each patch using the point map predicted in each frame via a weighted average.
Each token then calculates the 3D Euclidean distance to its eight neighboring
tokens in the image plane, with the average distance defined as $d_{i}= \mathrm{0.125}
\sum_{j \in \mathcal{N}(i)}\|P_{i}- P_{j}\|_{2}$, where $P_{i}$ is the 3D
position of token $i$, $\mathcal{N}(i)$ represents the set of its eight neighboring
tokens, and $\|\cdot\|_{2}$ is the Euclidean norm.

The optimal image voxel size $v_{\text{img}}$ is determined as the minimum
$d_{i}$ across all tokens to balance memory usage and storage efficiency. The scene
voxel size $v_{\text{scene}}$ is computed as the average of image voxel sizes across
all frames:
\begin{equation}
    v_{\text{img}}= \min_{i}d_{i}\quad \text{and}\quad v_{\text{scene}}= \frac{1}{t-1}
    \sum_{j=1}^{t-1}v_{\text{img},j},
\end{equation}
where $t$ denotes the sequence index of the current frame.
Given the model's streaming architecture, $v_{\text{scene}}$ undergoes
continuous online updates during inference, enabling adaptive scene-specific adjustments
across temporal sequences.

\paragraph{3D Spatial Memory Pruning.}
Once the scene voxel size is determined, tokens with similar 3D positions are grouped
into the same voxel, while tokens that are farther apart are assigned to
different voxels. The cumulative attention weight of each token is tracked, and
only the token with the highest weight within each voxel is retained, as illustrated
in \cref{fig:3dmem}. This mechanism effectively balances memory size to avoid
storing similar memories while preserving the spatial representation of the
scene.

\input{table/replica_table}

\input{fig/replica_vis}

\subsection{Training}
\label{train}
\paragraph{Loss Function.}
Following the approach in \cite{wang2024spann3r,wang2024dust3r}, we employ a confidence-aware
loss $\mathcal{L}_{\mathrm{conf}}$ for 3D regression and a scale loss $\mathcal{L}
_{\mathrm{scale}}$, which encourages the predicted point cloud to have an
average distance smaller than that of the ground truth. Overall, the final loss
function is:
\begin{equation}
    \mathcal{L}= \mathcal{L}_{\mathrm{conf}}+\mathcal{L}_{\mathrm{scale}}.
\end{equation}

\paragraph{Two-stage Curriculum Training.}
\label{curr_train} To enable our model to better handle long sequences, we adopt a two-stage
training strategy. In the first stage, the model is trained by randomly sampling
5 frames per video sequence.
This initial training phase allows the model to develop a preliminary
understanding, enabling the encoder to be frozen while fine-tuning subsequent modules
in the second stage. In the second stage, the ViT encoder remains frozen while the
other modules are fine-tuned, allowing the model to be trained with an increased
number of frames for addressing long sequences. Specifically, we initially sample 10
frames and subsequently 32 frames, enabling the model to gradually adapt to longer
sequences. This phased approach enhances long-sequence reconstruction
capabilities by exploiting spatio-temporal feature correlations across
progressively expanded temporal contexts, thereby optimizing the model's ability
to capture and utilize memory-related patterns in sequential data processing.

%% file: fig/3dmem_prune.tex
\begin{figure}[t]
    \centering
    \includegraphics[width=0.9\linewidth]{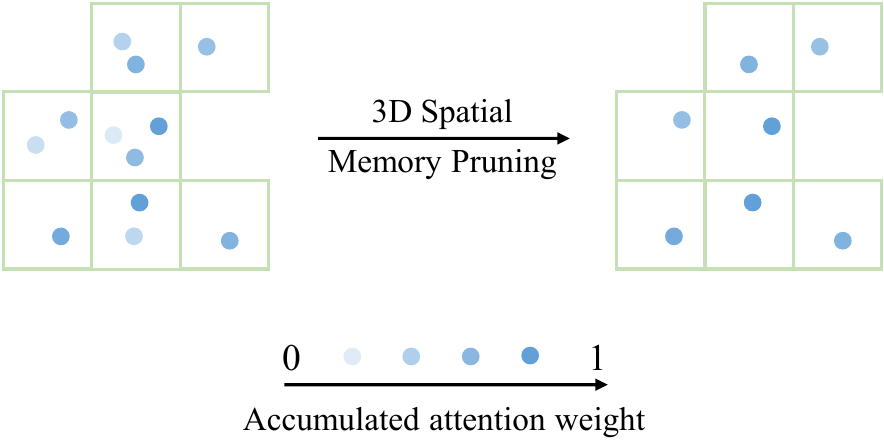}
    \caption{\textbf{3D Spatial Memory Pruning}. Memory tokens are grouped into
    voxels based on 3D positions (illustrated here in a 2D simplified view), with
    only the token having the highest accumulated attention weight retained per voxel.
    Darker blue dots indicate higher attention weights.}
    \label{fig:3dmem}
\end{figure}

%% file: table/7scenes_table.tex
\begin{table*}
    [t]
    \centering
    \footnotesize
    \setlength{\tabcolsep}{0.3em}
    \begin{tabularx}
        {\textwidth}{r >{\centering\arraybackslash}X >{\centering\arraybackslash}X >{\centering\arraybackslash}X >{\centering\arraybackslash}X >{\centering\arraybackslash}X >{\centering\arraybackslash}X >{\centering\arraybackslash}X >{\centering\arraybackslash}X > {\centering\arraybackslash}X >{\centering\arraybackslash}X > {\centering\arraybackslash}X > {\centering\arraybackslash}X >{\centering\arraybackslash}X}
        \toprule
        \multirow[b]{3}{*}{Method} & \multicolumn{6}{c}{7Scenes} & \multicolumn{6}{c}{NRGBD} & \multirow[b]{3}{*}{FPS}
        \\
        \cmidrule(lr){2-7} \cmidrule(lr){8-13} &
        \multicolumn{2}{c}{{Acc}$\downarrow$} &
        \multicolumn{2}{c}{{Comp}$\downarrow$} &
        \multicolumn{2}{c}{{NC}$\uparrow$} &
        \multicolumn{2}{c}{{Acc}$\downarrow$} &
        \multicolumn{2}{c}{{Comp}$\downarrow$} &
        \multicolumn{2}{c}{{NC}$\uparrow$} &
        \\
        \cmidrule(lr){2-3} \cmidrule(lr){4-5} \cmidrule(lr){6-7} \cmidrule(lr){8-9}
        \cmidrule(lr){10-11} \cmidrule(lr){12-13} & {Mean} & {Med.} & {Mean} & {Med.}
        & {Mean} & {Med.} & {Mean} & {Med.} & {Mean} & {Med.} & {Mean} & {Med.}
        & \\ \midrule
        \textbf{F-Recon~\cite{xu2023frozenrecon}} & 12.43 & 7.62 & 5.54 & \bf 2.31
        & \bf 61.89 & \bf 68.85 & 28.55 & 20.59 & 15.05
        & 6.31 & \bf 65.47 & \bf 75.77 & \cellcolor{color8!25} $\ll$1 \\
        \textbf{Dust3R~\cite{wang2024dust3r}} & 3.01 & \bf 1.47 & \bf
        5.11 & 2.79 & 58.83 & 63.73 & 3.94 & \bf 2.48 & \bf 5.31 & \bf
        3.58 & 62.62 & 72.29 & \cellcolor{color5!25} $\leq$3 \\
        \textbf{MASt3R~\cite{leroy2024mast3r}} & \bf2.82 & 1.56 &
        5.26 & 3.24 & 58.22 & 62.46 & \bf 3.85 & 2.54 & 5.50 & 3.62 &
        60.92 & 68.67 & \cellcolor{color5!25} $\leq$3 \\
        \midrule \textbf{MV-DUSt3R~\cite{tang2024mvdust3r}} & \bf 2.92 & 1.24 &
        \bf 2.49 & \bf 0.78 & \bf 66.42 & 76.07 & 3.76 & 1.99 & \bf
        2.55 & 0.92 & 81.16 & 95.39 & \cellcolor{color3!25} $\sim$15 \\ 
        \textbf{MV-DUSt3R+~\cite{tang2024mvdust3r}} & 2.93 & \bf 1.07 & 8.63 & 0.95
        & 66.38 & \bf 76.18 & \bf 3.47 & \bf 1.60 & 3.69 & \bf 0.85 & \bf
        84.33 & \bf 97.27 & \cellcolor{color5!25} $\sim$3 \\ 
        \midrule \rowcolor{gray!15} \textbf{CUT3R~\cite{wang2025cut3r}}
        & 7.73 & 3.57 & 7.75 & 1.83 & 65.74 & 73.98 & 12.48 & 5.57 &
        6.34 & 2.35 & 75.84 & 90.05 & \cellcolor{color1!25} $\sim$23 \\ 
        \rowcolor{gray!15} \textbf{Spann3R\cite{wang2024spann3r}} & 3.42 & 1.48
        & 2.41 & 0.85 & 66.35 & 76.25 & 6.91 & 3.15 & \bf 2.91 & \bf
        1.10 & \bf 77.75 & \bf 93.71 & \cellcolor{color2!25} $\sim$22 \\
        \rowcolor{gray!15} \textbf{Ours} & \bf 2.57 & \bf 1.14 & \bf 2.08 & \bf
        0.73 & \bf 66.55 & \bf 76.43 & \bf 6.66 & \bf 2.54 & 3.11 & 1.21
        & 77.56 & 93.08 & \cellcolor{color2!25} $\sim$22 \\ \bottomrule
    \end{tabularx}%
    \vspace{-5pt}
    \caption{\textbf{Quantitative results on 7Scenes~\cite{shotton2013scene} and
    NRGBD~\cite{azinovic2022neural} datasets.} All models are using $224\times 224$ image inputs.
    }
    \vspace{-10pt}
    \label{tab:main_table}
\end{table*}

%% file: fig/7scenes_vis.tex
\begin{figure*}[t]
    \centering
    \includegraphics[width=0.95\textwidth]{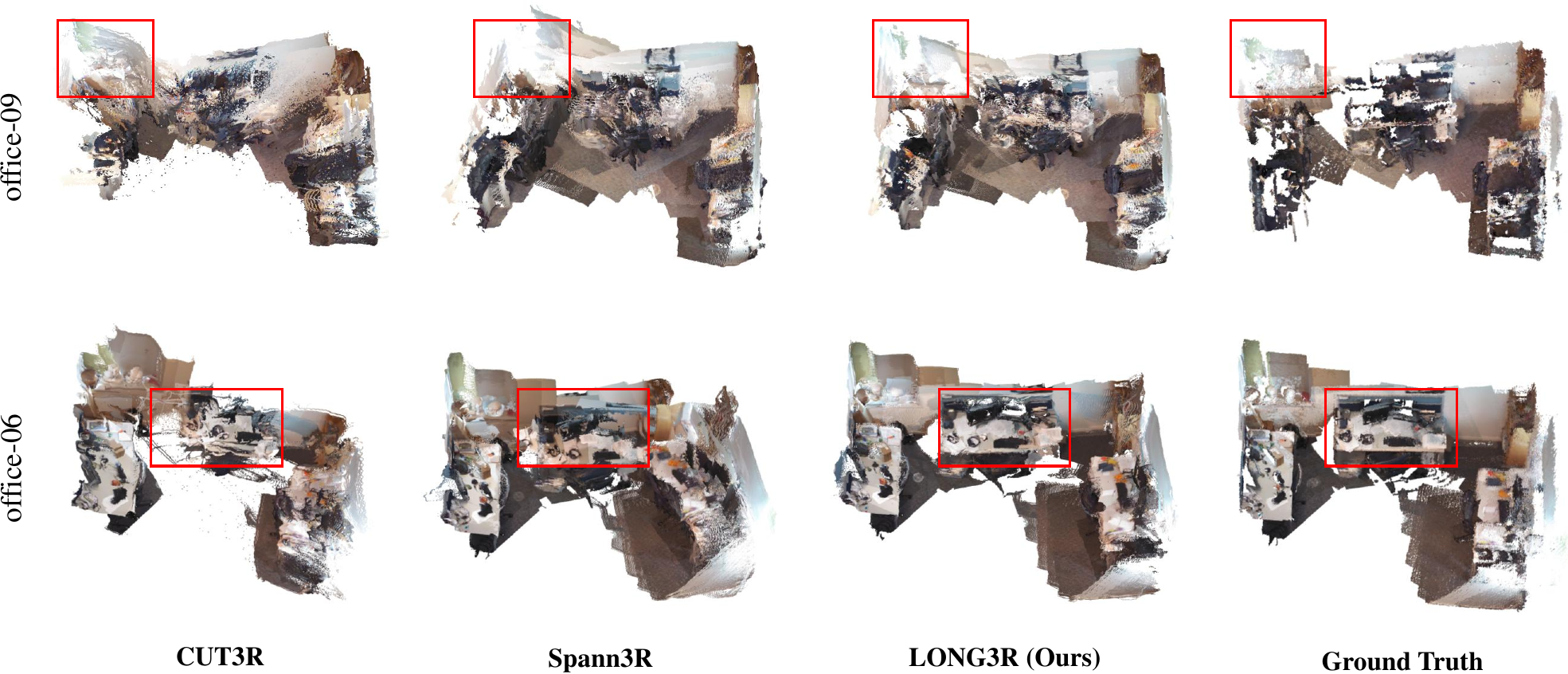}
    \caption{\textbf{Qualitative comparisons.} We present a comparison of reconstruction
    results on Office-06 and Office-09 from the 7Scenes~\cite{shotton2013scene} dataset
    with the Spann3R~\cite{wang2024spann3r} and CUT3R~\cite{wang2025cut3r} methods.
    In comparison with other online reconstruction methods, our approach achieves
    superior spatial consistency (e.g., the regions enclosed by the red bounding boxes) while preserving real-time performance.}
    \label{fig:7scenes_vis}
\end{figure*}

%% file: table/replica_table.tex
\begin{table*}
    [t]
    \centering
    \footnotesize
    \setlength{\tabcolsep}{0.3em}
    \begin{tabularx}
        {\textwidth}{r >{\centering\arraybackslash}X >{\centering\arraybackslash}X >{\centering\arraybackslash}X >{\centering\arraybackslash}X >{\centering\arraybackslash}X >{\centering\arraybackslash}X >{\centering\arraybackslash}X >{\centering\arraybackslash}X >{\centering\arraybackslash}X >{\centering\arraybackslash}X >{\centering\arraybackslash}X >{\centering\arraybackslash}X >{\centering\arraybackslash}X}
        \toprule
        \multirow[b]{3}{*}{Method} & \multicolumn{6}{c}{Replica$_{\mathrm{100}}$}
        & \multicolumn{6}{c}{Replica$_{\mathrm{200}}$} & \multirow[b]{3}{*}{{FPS}} \\
        \cmidrule(lr){2-7} \cmidrule(lr){8-13} &
        \multicolumn{2}{c}{{Acc}$\downarrow$} &
        \multicolumn{2}{c}{{Comp}$\downarrow$} &
        \multicolumn{2}{c}{{NC}$\uparrow$} &
        \multicolumn{2}{c}{{Acc}$\downarrow$} &
        \multicolumn{2}{c}{{Comp}$\downarrow$} &
        \multicolumn{2}{c}{{NC}$\uparrow$} & \\
        \cmidrule(lr){2-3} \cmidrule(lr){4-5} \cmidrule(lr){6-7} \cmidrule(lr){8-9}
        \cmidrule(lr){10-11} \cmidrule(lr){12-13} & {Mean} & {Med.} & {Mean} &
        {Med.} & {Mean} & {Med.} & {Mean} & {Med.} & {Mean} & {Med.} & {Mean}
        & {Med.} & \\ \midrule
        \textbf{{Dust3R}~\cite{wang2024dust3r}} & 6.34 & 3.99 & 6.44
        & 3.68 & 61.67 & 69.27 & \bf 4.99 & \bf 2.76 & \bf 4.63 & \bf
        2.59 & \bf 62.26 & \bf 70.76 & \cellcolor{color5!25} $\leq$3 \\ 
        \textbf{{MASt3R}~\cite{leroy2024mast3r}} & \bf 5.10 & \bf
        2.96 & \bf 6.00 & \bf 3.43 & \bf 61.81 & \bf 69.52 & 5.26 & 3.23
        & 7.31 & 3.75 & 58.03 & 62.81 & \cellcolor{color5!25} $\leq$3 \\ \midrule 
        \textbf{{MV-DUSt3R}~\cite{tang2024mvdust3r}} & 10.41 & 6.48 & 4.34 & 1.22
        & 73.76 & 88.36 & 17.02 & 11.70 & \bf 5.10 & \bf 1.36 & 66.74 & 78.24
        & \cellcolor{color4!25} $\sim$7 \\ 
        \textbf{{MV-DUSt3R+}~\cite{tang2024mvdust3r}} & \bf 5.28 & \bf 3.26 &
        \bf \bf 2.56 & \bf 0.89 & \bf 79.07 & \bf 93.63 & \bf 11.79 & \bf
        8.37 & 5.64 & 1.53 & \bf 70.66 & \bf 83.86 & \cellcolor{color7!25} $\sim$1 \\ 
        \midrule \rowcolor{gray!15} \textbf{{CUT3R}~\cite{wang2025cut3r}}
        & 20.44 & 14.64 & 5.67 & 2.32 & 69.63 & 84.31 & 28.3 & 20.68
        & 6.61 & 1.88 & 63.95 & 73.85 & \cellcolor{color1!25} $\sim$23\\ 
        \rowcolor{gray!15} \textbf{Spann3R\cite{wang2024spann3r}} & 14.08 & 8.88
        & 4.67 & 1.61 & 72.46 & 88.98 & 16.29 & 10.17 & 4.02 & 1.16 &
        68.56 & 82.80 & \cellcolor{color2!25} $\sim$21\\ 
        \rowcolor{gray!15} \textbf{Ours} & \bf 11.46 & \bf 7.55 & \bf 3.68 & \bf
        1.24 & \bf 73.29 & \bf 89.86 & \bf 11.93 & \bf 7.42 & \bf 2.73 &
        \bf 0.87 & \bf 68.67 & \bf 82.92 & \cellcolor{color2!25} $\sim$21\\ \bottomrule
    \end{tabularx}%
    \vspace{-5pt}
    \caption{\textbf{Quantitative results on Replica~\cite{replica19arxiv}
    datasets.} 
    All models are using $224\times 224$ image inputs. 
    Replica$_{\mathrm{100}}$ and Replica$_{\mathrm{200}}$ represent sequence lengths of 100 and 200 frames, respectively.}
    \vspace{-10pt}
    \label{tab:replica_table}
\end{table*}

%% file: fig/replica_vis.tex
\begin{figure*}[t]
    \centering
    \includegraphics[width=0.95\textwidth]{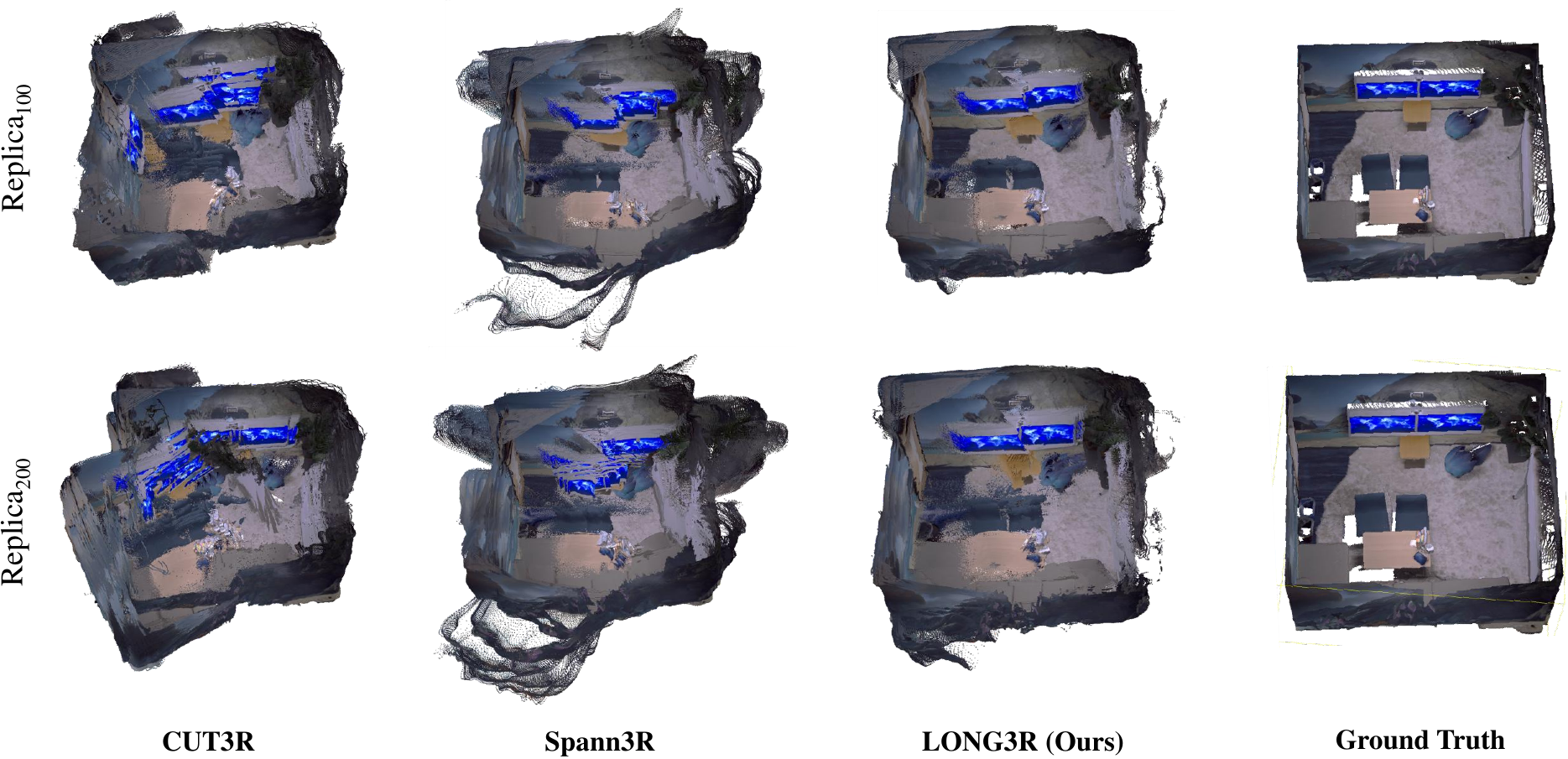}
    \caption{\textbf{Qualitative comparisons.} We present a comparison of reconstruction
    results on Office-0 from the Replica~\cite{replica19arxiv} dataset with the Spann3R~\cite{wang2024spann3r}
    and CUT3R~\cite{wang2025cut3r} methods. During streaming observations within the same scene, existing methods suffer from increasing ambiguity
    due to error accumulation. In contrast, our approach effectively suppresses cumulative
    drift and maintains spatial consistency.}
    \label{fig:replica_vis}
\end{figure*}

%% file: sec/4_experiment.tex
\section{Experiments}
To comprehensively assess overall performance and component effectiveness, we evaluate our method on 3D reconstruction (\cref{eval_3d_recon}) and camera pose estimation (\cref{eval_cam_pose}), along with ablation and analysis presented in \cref{ablation}.

\subsection{Setup}

\paragraph{Training Datasets.}
Following Spann3R~\cite{wang2024spann3r}, we train our model with a mixture of 6 datasets:
Habitat~\cite{savva2019habitat}, ARKitScenes~\cite{baruch2021arkitscenes}, BlendedMVS~\cite{yao2020blendedmvs},
ScanNet++~\cite{yeshwanth2023scannet++}, Co3D-v2~\cite{reizenstein2021common}, ScanNet~\cite{dai2017scannet}.
These datasets integrate real-world and synthetic data, encompassing metric-scale
measurements and normalized-scale samples.

\paragraph{Baselines.}
For online streaming reconstruction methods, we evaluate \ourwork with Spann3R~\cite{wang2024spann3r}
and CUT3R~\cite{wang2025cut3r} as primary baselines. To ensure a comprehensive analysis,
we further compare our method with DUSt3R~\cite{wang2024dust3r} and MASt3R~\cite{leroy2024mast3r},
which involve additional post-processing stages, as well as offline approaches
MV-DUSt3R and MV-DUSt3R+~\cite{tang2024mvdust3r}. All evaluations are conducted
on a single NVIDIA RTX 3090 GPU with 24 GB of VRAM, and all input images are resized
to $224\times 224$ for a fair comparison.

\paragraph{Implementation details.}
We use ViT-Large~\cite{dosovitskiy2020image} as \ourwork's encoder, initialized with
DUSt3R's encoder weights. Both training stages utilize images at a resolution of
$224\times 224$. 
Experiments are conducted with 10-frame short-term memory and 3000-token long-term memory.
In the first stage, we employ the AdamW optimizer with a
learning rate of $1.12 \times 10^{-4}$ and a batch size of $10$ per GPU, training
for $120$ epochs.
In the second stage, we fine-tune the model using the AdamW optimizer with a learning
rate of $1 \times 10^{-5}$ on 10-view and 32-view sequences, training for $12$
epochs each. The first stage runs on $16$ A100 GPUs for $28$ hours, while fine-tuning
requires about $20$ hours on the same hardware configuration.

\input{table/pose_est_table}

\subsection{3D Reconstruction}
\label{eval_3d_recon}
We evaluate scene-level reconstruction on three unseen datasets: 7Scenes~\cite{shotton2013scene},
NRGBD~\cite{azinovic2022neural}, and Replica~\cite{replica19arxiv} to
demonstrate generalization capability in long-sequence reconstruction using \textit{accuracy}, \textit{completion} and \textit{normal consistency} as
in previous works~\cite{wang2024spann3r, azinovic2022neural, wang2023co, zhu2022nice} and we report accuracy and completeness in centimeters. For the Replica~\cite{replica19arxiv}
dataset, we consider two settings: uniformly sampling 100 and 200 frames,
denoted Replica$_{\mathrm{100}}$ and Replica$_{\mathrm{200}}$, respectively.
\paragraph{Quantitative comparisons.}
\cref{tab:main_table} and \cref{tab:replica_table} present the quantitative comparison
of reconstruction metrics between our method and existing 3D reconstruction approaches.
The learning-based approaches evaluated are categorized into three classes: optimization
methods with post-processing refinement like DUSt3R~\cite{wang2024dust3r} and MASt3R~\cite{leroy2024mast3r},
offline methods like MV-DUSt3R~\cite{tang2024mvdust3r}, and streaming online reconstruction
methods like Spann3R~\cite{wang2024spann3r} and CUT3R~\cite{wang2025cut3r}. As
demonstrated in \cref{tab:main_table}, our method achieves the highest reconstruction
accuracy compared to online streaming reconstruction methods, while attaining a
precision comparable to post-optimization and offline approaches on the 7Scenes~\cite{shotton2013scene}
and NRGBD~\cite{azinovic2022neural} datasets. Furthermore, our real-time performance offers a crucial advantage over the slower post-optimization and offline methods.

To systematically assess the robustness and geometric fidelity of reconstruction
approaches under streaming observations, we design a controlled experiment on the
Replica~\cite{replica19arxiv} dataset. We sample $\mathrm{100}$ and
$\mathrm{200}$ frames sequentially from the same scene as complete observations.
The predicted pointmaps are aggregated as the reconstruction results for metric computation,
reflecting the impact of sequence length on reconstruction accuracy. As shown in~\cref{tab:replica_table}, post-optimization methods achieve higher accuracy due to global alignment effectively mitigating drift, albeit at the cost of significantly lower FPS, which limits their applicability in real-time scenarios.
Other optimization-free baselines exhibit reduced robustness with denser observational inputs due to their
inherent lack of holistic spatial consistency awareness during training and inference.
In contrast, our approach maintains competitive reconstruction accuracy compared
to online methods while avoiding catastrophic performance degradation on extended
sequences.

\input{table/ablation_memory_gating}
\input{fig/mem_gating}

\paragraph{Qualitative comparisons.}
We qualitatively compare our method with online reconstruction baselines on the
7Scenes~\cite{shotton2013scene} and Replica~\cite{replica19arxiv} datasets. Detailed
comparisons of reconstruction results are presented in \cref{fig:7scenes_vis} and
\cref{fig:replica_vis}. As shown in \cref{fig:7scenes_vis}, our method achieves superior
spatial consistency while maintaining real-time performance compared to Spann3R~\cite{wang2024spann3r}
and CUT3R~\cite{wang2025cut3r}. \cref{fig:replica_vis} presents results from
long-sequence streaming observation experiments. With increasing observation sequences,
Spann3R~\cite{wang2024spann3r} and CUT3R~\cite{wang2025cut3r} exhibit amplified
ambiguity in geometric predictions (e.g., erroneous surfaces of televisions), whereas
our approach maintains consistent reconstruction quality via 3D spatiotemporal context
aggregation.

\subsection{Camera Pose Estimation}
\label{eval_cam_pose}
We evaluate camera pose estimation on 7Scenes~\cite{shotton2013scene}, TUM~\cite{tum_dynamic}, and ScanNet~\cite{dai2017scannet} using Absolute Translation Error (ATE), Relative Translation Error (RPE$_t$), and Relative Rotation Error (RPE$_r$), following ~\cite{chen2024leap, zhang2024monst3r, zhao2022particlesfm}.

As shown in~\cref{tab:quant_pose}, our method significantly outperforms Spann3R and CUT3R on static scene datasets, namely 7Scenes and ScanNet. Despite being trained exclusively on static scenes, our method remains competitive with Spann3R and CUT3R on the TUM Dynamics dataset, which features dynamic human motion.

\subsection{Ablation and analysis}
\label{ablation}

\paragraph{Attention-based Memory Gating.}
We analyze the impact of the memory gating mechanism on the 7Scenes~\cite{shotton2013scene} and NRGBD~\cite{azinovic2022neural} datasets. The experimental results are summarized in \cref{tab:ab_gating}, and \cref{fig:mem_num}. The memory gating mechanism removes memory features irrelevant to the current frame, exemplified by a 27\% reduction on 7Scenes, and achieves an optimal balance between reconstruction accuracy and computational efficiency in streaming reconstruction. We evaluated the FPS with (21.4 FPS) and without (18.0 FPS) memory gating, resulting in a 20\% boost.

\input{fig/ablation_mem_decoder}

\paragraph{Dual-Source Refined Decoder.}
We conduct an ablation study on the Replica~\cite{replica19arxiv} dataset to compare the effectiveness of various dual-source block designs, with \cref{fig:decoder_ablation} depicting the design variations.
Due to memory constraints associated with the concatenation method, the experiments reported in \cref{tab:ab_concat} use a sequence length of 24 frames instead of 32 during the second stage of training. As shown in \cref{tab:ab_concat}, compared with the concatenation approach, our proposed interleaved attention blocks yield superior scene reconstruction accuracy under both sampling settings in Replica while reducing computational complexity. These performance gains primarily stem from mitigating information loss caused by feature space misalignment between memory features and the coarse features of frame $t + 1$. Our interleaved attention blocks address this issue by employing alternating cross-attention, which progressively aligns feature spaces and improves computational efficiency.

\paragraph{3D Spatio-Temporal Memory.}
We conduct an ablation study in two parts to evaluate our 3D spatio-temporal memory design. In the first part, we compare the complete design with a variant that excludes the long-term 3D spatial memory component on the Replica~\cite{replica19arxiv} dataset. The results detailed in \cref{tab:ab_memory} reveal that omitting the long-term 3D spatial memory significantly degrades reconstruction performance, highlighting its essential role in long-sequence streaming reconstruction. In the second part, we compare our proposed design with the Spann3R memory~\cite{wang2024spann3r} framework under consistent network architectures. Our findings demonstrate that our 3D spatio-temporal memory achieves superior scene-level reconstruction accuracy through reducing spatially redundant memory while preserving coherent geometric understanding across sequential frames.

\input{table/ablation_memory_decoder}

\input{table/ablation_memory_system.tex}

%% file: table/pose_est_table.tex
\begin{table}[t]
  \centering
  \footnotesize
  \setlength{\tabcolsep}{-0.2em}
    \begin{tabularx}{\columnwidth}{p{1.8cm} >{\centering\arraybackslash}X >{\centering\arraybackslash}X >{\centering\arraybackslash}X >{\centering\arraybackslash}X >{\centering\arraybackslash}X >{\centering\arraybackslash}X >{\centering\arraybackslash}X >{\centering\arraybackslash}X >{\centering\arraybackslash}X}
      \toprule
         \multirow{2}{*}{Method} & \multicolumn{3}{c}{{7Scenes}} & \multicolumn{3}{c}{{TUM}} & \multicolumn{3}{c}{{ScanNet}} \\ 
      \cmidrule(lr){2-4} \cmidrule(lr){5-7} \cmidrule(lr){8-10}
         & ATE & RPE$_t$ & RPE$_r$ & ATE & RPE$_t$ & RPE$_r$  & ATE & RPE$_t$ & RPE$_r$ \\
      \midrule
        \textbf{Spann3R~\cite{wang2024spann3r}} 
        & 12.64 & 6.15 & 1.88 & 5.66 & \bf2.13 & \bf0.59 & 9.83 & 2.30 & 0.66 \\
        \textbf{CUT3R~\cite{wang2025cut3r}} 
        & 12.40 & 7.65 & 2.34 & 6.25 & 2.55 & 0.69 & 14.27 & 3.58 & 0.92 \\
        \textbf{Ours} 
        & \bf8.72 & \bf5.03 & \bf1.67 & \bf5.40 & 2.36 & 0.60 & \bf6.44 & \bf2.14 & \bf0.61 \\
      \bottomrule
    \end{tabularx}
    \vspace{-5pt}
    \caption{\textbf{Evaluation of Camera Pose Estimation (cm/$^\circ$).} All three models using 224 × 224 image inputs.} 
    \vspace{-1.2em}
    \label{tab:quant_pose}
\end{table}

%% file: table/ablation_memory_gating.tex
\begin{table}[t]
    \centering
    \footnotesize
    \setlength{\tabcolsep}{0.2em}
    \begin{tabularx}
        {\columnwidth}{r >{\centering\arraybackslash}X >{\centering\arraybackslash}X >{\centering\arraybackslash}X >{\centering\arraybackslash}X >{\centering\arraybackslash}X >{\centering\arraybackslash}X >{\centering\arraybackslash}X >{\centering\arraybackslash}X >{\centering\arraybackslash}X}
        \toprule
        \multirow[b]{3}{*}{Method} & \multicolumn{4}{c}{7Scenes} & \multicolumn{4}{c}{NRGBD} & \multirow[b]{3}{*}{FPS}
        \\ \cmidrule(lr){2-5} \cmidrule(lr){6-9} & \multicolumn{2}{c}{{Acc}$\downarrow$}
        & \multicolumn{2}{c}{{Comp}$\downarrow$} & \multicolumn{2}{c}{{Acc}$\downarrow$}
        & \multicolumn{2}{c}{{Comp}$\downarrow$} & \\ \cmidrule(lr){2-3} \cmidrule(lr){4-5}
        \cmidrule(lr){6-7} \cmidrule(lr){8-9} & {Mean} & {Med.} & {Mean} & {Med.}
        & {Mean} & {Med.} & {Mean} & {Med.} & \\ \midrule \textbf{w/o Gating}
        & \bf 2.53 & \bf 1.12 & 2.12 & 0.74 & 6.72 & 3.14 & \bf 2.91 & \bf 1.20 & 18.0
        \\ \textbf{w/ Gating} & 2.57 & 1.14 & \bf 2.08 & \bf 0.73 & \bf 6.66
        & \bf 3.11 & 2.92 & 1.21 & \bf 21.4 \\ \bottomrule
    \end{tabularx}%
    \vspace{-5pt}
    \caption{\textbf{Attention-based Memory Gating ablation study on 7Scenes~\cite{shotton2013scene}
    and NRGBD~\cite{azinovic2022neural}.}}
    \vspace{-5pt}
    \label{tab:ab_gating}
\end{table}

%% file: fig/mem_gating.tex
\begin{figure}[t]
    \centering
    \includegraphics[width=0.9\linewidth]{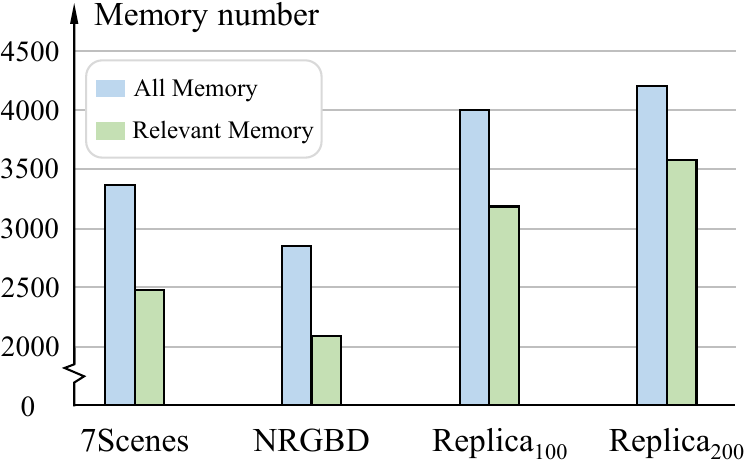}
    \caption{\textbf{Memory number comparison.} 
    This figure illustrates the number of memory tokens before (all memory) and after (relevant memory) passing through the memory gating module.}
    \label{fig:mem_num}
\end{figure}

%% file: fig/ablation_mem_decoder.tex
\begin{figure}[t]
    \centering
    \includegraphics[width=0.9\linewidth]{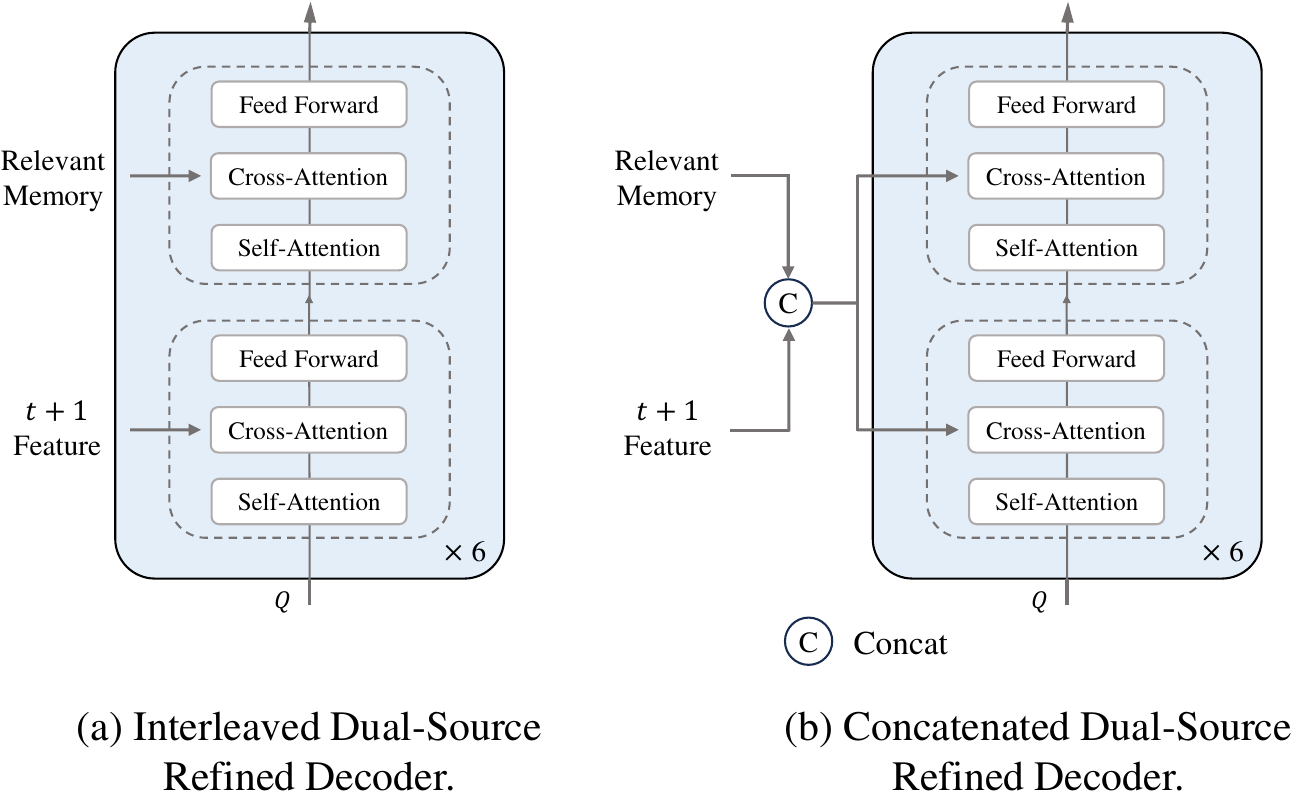}
    \caption{\textbf{Dual-Source Decoder Comparison.} This figure compares the interleaved and concatenated architectures of the Dual-Source Refined Decoder.}
    \label{fig:decoder_ablation}
\end{figure}

%% file: table/ablation_memory_decoder.tex
\begin{table}[t]
    \centering
    \footnotesize
    \setlength{\tabcolsep}{0.2em}
    \begin{tabularx}
        {\columnwidth}{r >{\centering\arraybackslash}X >{\centering\arraybackslash}X >{\centering\arraybackslash}X >{\centering\arraybackslash}X >{\centering\arraybackslash}X >{\centering\arraybackslash}X >{\centering\arraybackslash}X >{\centering\arraybackslash}X}
        \toprule

        \multirow[b]{3}{*}{Method} & \multicolumn{4}{c}{Replica$_{\mathrm{100}}$} & \multicolumn{4}{c}{Replica$_{\mathrm{200}}$}
        \\ \cmidrule(lr){2-5} \cmidrule(lr){6-9} & \multicolumn{2}{c}{{Acc}$\downarrow$}
        & \multicolumn{2}{c}{{Comp}$\downarrow$} & \multicolumn{2}{c}{{Acc}$\downarrow$}
        & \multicolumn{2}{c}{{Comp}$\downarrow$} \\ \cmidrule(lr){2-3} \cmidrule(lr){4-5}
        \cmidrule(lr){6-7} \cmidrule(lr){8-9} & {Mean} & {Med.} & {Mean} & {Med.}
        & {Mean} & {Med.} & {Mean} & {Med.} \\ \midrule \textbf{Concat.} & 14.83 & 10.26 & 4.59 & 1.81 & 29.52 & 21.04 & 8.88 & 4.05 \\ \textbf{Interleaved.}
        & \bf 12.06 & \bf 7.67 & \bf 3.68 & \bf 1.23 & \bf 13.34 & \bf 8.41 & \bf 3.15 &
        \bf 0.94 \\ \bottomrule
    \end{tabularx}%
    \vspace{-5pt}

    \caption{\textbf{Refined Decoder ablation study on Replica~\cite{replica19arxiv}}.}
    \vspace{-5pt}
    \label{tab:ab_concat}
\end{table}

%% file: table/ablation_memory_system.tex
\begin{table}[t]
    \centering
    \footnotesize
    \setlength{\tabcolsep}{0.2em}
    \begin{tabularx}
        {\columnwidth}{r >{\centering\arraybackslash}X >{\centering\arraybackslash}X >{\centering\arraybackslash}X >{\centering\arraybackslash}X >{\centering\arraybackslash}X >{\centering\arraybackslash}X >{\centering\arraybackslash}X >{\centering\arraybackslash}X}
        \toprule

        \multirow[b]{3}{*}{Method} & \multicolumn{4}{c}{7Scenes} & \multicolumn{4}{c}{Replica$_{\mathrm{200}}$}
        \\ \cmidrule(lr){2-5} \cmidrule(lr){6-9} & \multicolumn{2}{c}{{Acc}$\downarrow$}
        & \multicolumn{2}{c}{{Comp}$\downarrow$} & \multicolumn{2}{c}{{Acc}$\downarrow$}
        & \multicolumn{2}{c}{{Comp}$\downarrow$} \\ \cmidrule(lr){2-3} \cmidrule(lr){4-5}
        \cmidrule(lr){6-7} \cmidrule(lr){8-9} & {Mean} & {Med.} & {Mean} & {Med.}
        & {Mean} & {Med.} & {Mean} & {Med.} \\ \midrule \textbf{w/o 3D Spa. Mem.}
        & 5.76 & 2.96 & 3.30 & 1.22 & 65.75 & 47.63 & 13.24 & 3.49 \\ \textbf{w/
        Spann3R Mem.} & 2.64 & 1.16 & 2.10 & 0.74 & 12.41 & 7.87 & 3.07 & 0.88 \\
        \midrule \textbf{\ourwork (ours)} & \bf 2.57 & \bf 1.14 & \bf 2.08 & \bf 0.73 & \bf
        11.93 & \bf 7.42 & \bf 2.74 & \bf 0.87 \\\bottomrule
    \end{tabularx}%
    \vspace{-5pt}

    \caption{\textbf{Memory frameworks ablation study on 7Scenes~\cite{shotton2013scene}
    and Replica~\cite{replica19arxiv}.} w/o 3D Spa. Mem.: without 3D spatial memory, only with temporal memory.}
    \vspace{-5pt}
    \label{tab:ab_memory}
\end{table}

%% file: sec/5_conclusion.tex
\section{Conclusion}

In this paper, we propose \ourwork, a novel framework for long-sequence streaming 3D reconstruction that overcomes key limitations of existing methods. It combines a memory gating mechanism, a Dual-Source Refined Decoder, and a dynamic 3D Spatio-Temporal Memory Module to improve memory efficiency and reduce redundancy. A two-stage curriculum training strategy further helps the model adapt to long sequences. Experiments on multiple datasets demonstrate that \ourwork achieves state-of-the-art while maintaining real-time performance. Future work will explore scaling the framework to more complex environments.

\noindent
\textbf{Limitations.} Since our predictions are defined relative to the first frame, our model may produce blurry results if the viewpoint deviates significantly. 
Due to the lack of dynamic training data, the current model struggles to handle highly dynamic scenes with large object motions.

%% file: sec/6_acknowledgment.tex
\section*{Acknowledgements}
This work is supported by the National Key R\&D Program of China (2022ZD0161700) and Tsinghua University Initiative Scientific Research Program.